\pdfoutput=1

\documentclass[11pt]{article}

\usepackage[final]{acl}

\usepackage{times}
\usepackage{latexsym}
\usepackage{authblk}
\usepackage{hyperref}
\usepackage{CJKutf8}
\usepackage{times}
\usepackage{graphicx}
\usepackage{amsmath}
\usepackage{booktabs} 
\usepackage{multirow} 
\usepackage{multicol} 
\usepackage{subfig}
\usepackage{CJKutf8}
\usepackage{tabularx}
\usepackage{algorithm}
\usepackage{algpseudocode}
\usepackage{amsmath}

\usepackage[T1]{fontenc}

\usepackage[utf8]{inputenc}

\usepackage{microtype}

\usepackage{inconsolata}

\usepackage{graphicx}
\usepackage{svg}

\title{A Multilingual Dataset and Empirical Validation for the Mutual Reinforcement Effect in Information Extraction}

\author{
  \bf Chengguang Gan\textsuperscript{1}\thanks{\href{mailto:ganchengguan@yahoo.co.jp}{ganchengguan@yahoo.co.jp}},
  Sunbowen Lee\textsuperscript{2},
  Qingyu Yin\textsuperscript{3},
  Yunhao Liang\textsuperscript{4},
  Xinyang He\textsuperscript{5}, \\
  \bf
  Hanjun Wei\textsuperscript{4},
  Younghun Lim\textsuperscript{1},
  Shijian Wang\textsuperscript{6},
  Hexiang Huang\textsuperscript{7}, \\
  \bf
  Qinghao Zhang\textsuperscript{8},
  Shiwen Ni\textsuperscript{9}\textsuperscript{$\dagger$},
  Tatsunori Mori\textsuperscript{1}\textsuperscript{$\dagger$} \\
  \small{\textsuperscript{1}Yokohama National University,
  \textsuperscript{2}Shenzhen University of Advanced Technology,} \\
  \small{\textsuperscript{3}Zhejiang University,
  \textsuperscript{4}University of Chinese Academy of Sciences,} \\
  \small{\textsuperscript{5}Chengdu Institute of Computer Applications, Chinese Academy of Sciences,
  \textsuperscript{6}Southeast University,
  \textsuperscript{7}University of Tsukuba,} \\
  \small{\textsuperscript{8}Pusan National University,
  \textsuperscript{9}Shenzhen Institute of Advanced Technology, Chinese Academy of Sciences}
}

\begin{document}
\begin{CJK}{UTF8}{gbsn}

\maketitle

\begingroup
\renewcommand{\thefootnote}{\fnsymbol{footnote}}
\footnotetext[2]{Corresponding authors.}
\endgroup

\begin{abstract}

The Mutual Reinforcement Effect (MRE) describes a phenomenon in information extraction where word-level and sentence-level tasks can mutually improve each other when jointly modeled. While prior work has reported MRE in Japanese, its generality across languages and task settings has not been empirically validated, largely due to the lack of multilingual MRE datasets. To address this limitation, we introduce the Multilingual MRE Mix dataset (MMM), which consists of 21 sub-datasets covering English, Japanese, and Chinese. We propose an LLM-assisted dataset translation and alignment framework that significantly reduces manual annotation effort while preserving the structural requirements of MRE tasks. Building on MMM, we adopt a unified input-output framework to train an open-domain information extraction model and conduct extensive empirical studies, including full fine-tuning ablations and the construction of knowledgeable verbalizers based on MRE-mix data. Experimental results show that 76 percent of the MMM sub-datasets consistently exhibit the Mutual Reinforcement Effect across languages. These findings provide systematic empirical validation of MRE in multilingual settings and demonstrate its practical value for information extraction. The OIELLM model and datasets is open-source in HuggingFace: \href{https://ganchengguang.github.io/MRE/}{GitHub Website}\footnote{\url{https://ganchengguang.github.io/MRE/}}

\end{abstract}

\section{Introduction}

\begin{figure}[!t]
\centering
\includegraphics[width=219 pt]{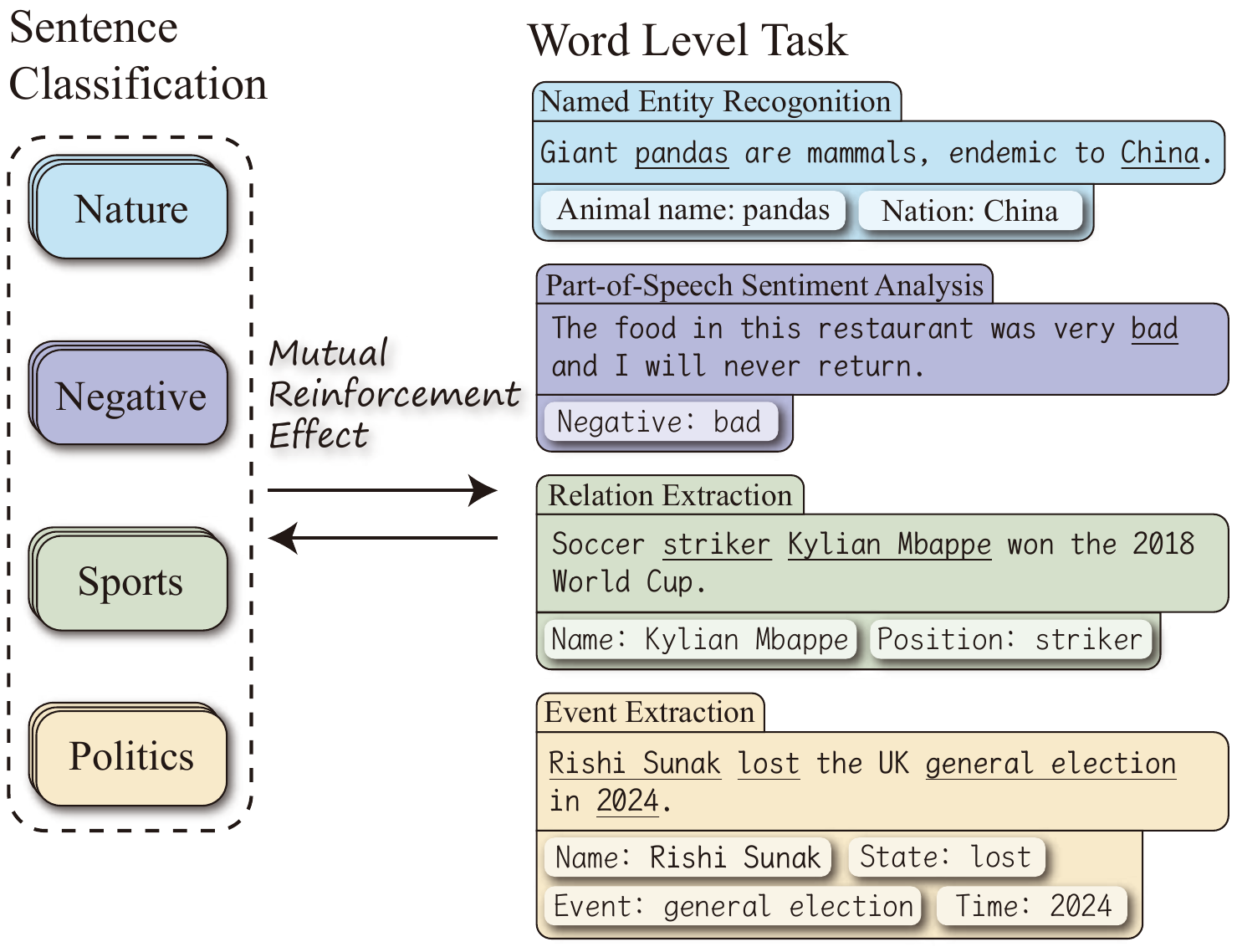}
\caption{\label{1figure1}The Mutual Reinforcement Effect between word-level labels and a text-level label within the same text. \textbf{A word-level IE task is regarded as a Point, and a text-level IE task is regarded as a Line. Mutual Reinforcement Effect exists between the Point and the Line.}}
\end{figure}

Information extraction (IE) \citet{sarawagi2008information} is a fundamental research area in natural language processing (NLP), aiming to transform unstructured text into structured representations. Over the years, IE has evolved into a collection of specialized subtasks, including sentence classification \citep{zhang2015sensitivity}, text classification \citep{lai2015recurrent}, Named Entity Recognition (NER) \citep{qu2023survey, nadeau2007survey, lample2016neural}, sentiment analysis \citep{tan2023survey, medhat2014sentiment, rodriguez2023review}, relation extraction \citep{wadhwa2023revisiting, mintz2009distant, etzioni2008open}, and event extraction \citep{gao2023exploring, 8918013}.  
Traditionally, these subtasks have been studied and modeled in isolation.

Multi-task learning for IE \citep{sun2023learning, zhao2020spanmlt} attempts to bridge this separation by jointly training multiple subtasks within a single model. In most existing approaches, datasets from different IE tasks are merged, and the model is optimized with task-specific output heads. While this paradigm allows models to share representations across tasks, it treats each task as an independent objective and does not explicitly model or analyze the semantic interactions between tasks. As a result, whether and how different IE subtasks can mutually benefit from each other remains underexplored.

The Mutual Reinforcement Effect (MRE) \citep{gan2023sentence} was proposed to explicitly study this interaction. MRE refers to the phenomenon that word-level IE tasks and text-level IE tasks can mutually enhance each other when they are jointly modeled on the same text. In this formulation, text-level tasks such as sentence classification or sentiment analysis provide global semantic context, while word-level tasks such as NER provide fine-grained semantic signals. Rather than treating them as parallel objectives, MRE emphasizes their bidirectional dependency within a single input instance.

To illustrate this idea, IE tasks are categorized into word-level tasks and text-level tasks, which are abstracted as Points and Lines, respectively. Understanding the Point facilitates understanding the Line, and vice versa. Figure~\ref{1figure1} presents a concrete example of this interaction. The sentence “Giant pandas are mammals, endemic to China.” is labeled as \textit{nature} at the text level, while containing word-level entity annotations such as \textit{Animal Name: pandas} and \textit{Nation: China}. The sentence-level label constrains the plausible entity types, while the recognized entities reinforce the semantic correctness of the sentence-level classification. This interaction is not task-specific but reflects a general mechanism of human language understanding, where global meaning is inferred from words and words are interpreted under global context \citep{gan2023think}.

Although MRE has been shown to be effective in prior studies, its empirical exploration has been severely constrained by data availability. Existing MRE mix datasets are limited to Japanese, which restricts systematic investigation across languages and hinders broader validation of the effect. As a consequence, it remains unclear whether MRE is language-specific or a more general property of IE tasks.

\begin{figure}[!t]
\centering
\includegraphics[width=200 pt]{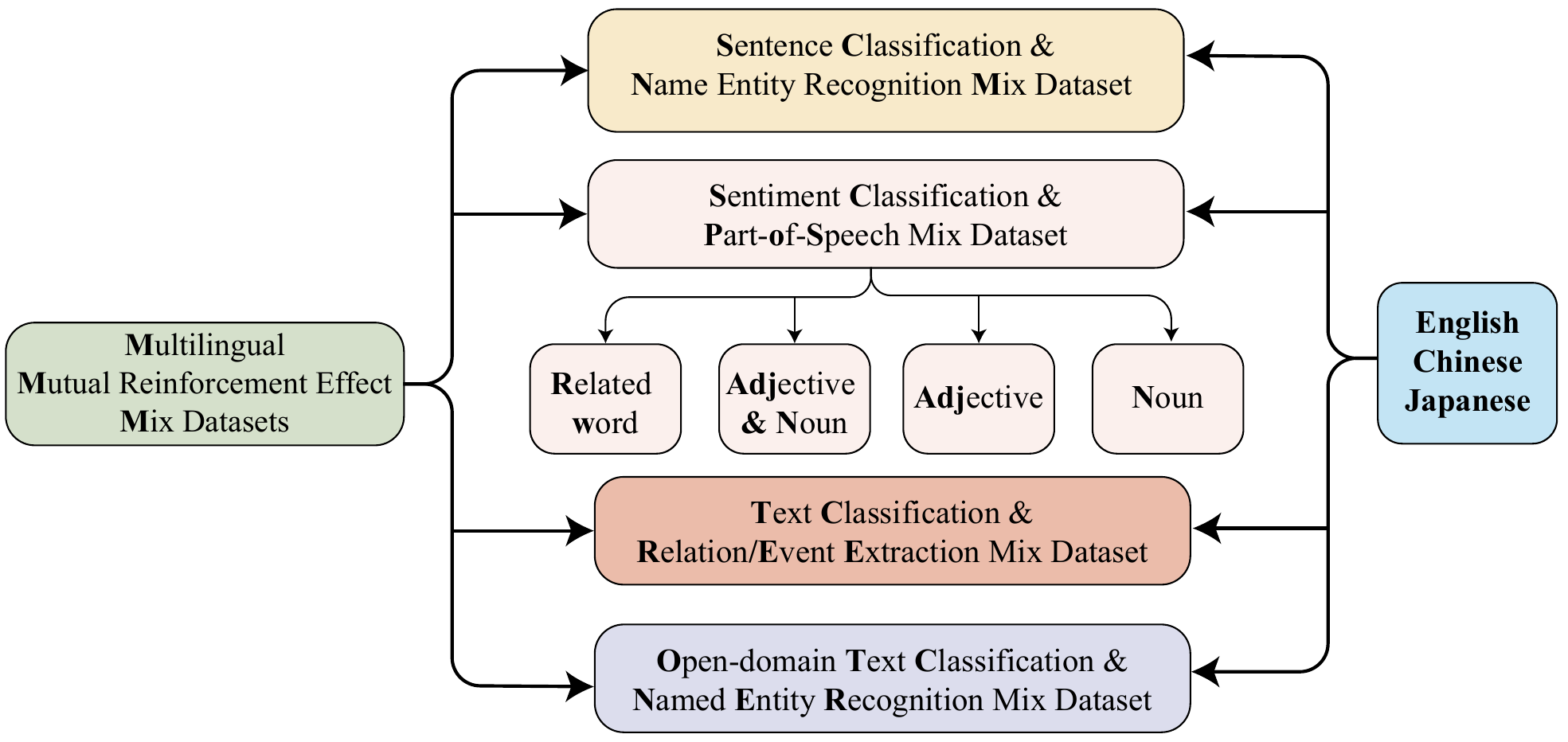}
\caption{\label{2figure2}Overview of the Multilingual Mutual Reinforcement Effect Mix (MMM) datasets and their sub-datasets.}
\end{figure}

To address this limitation, we construct the Multilingual Mutual Reinforcement Effect Mix (MMM) datasets, covering English, Chinese, and Japanese. As shown in Figure~\ref{2figure2}, MMM consists of 21 sub-datasets, with seven datasets per language. These datasets span multiple MRE task combinations, including sentence classification with NER (SCNM), sentiment classification with part-of-speech related supervision (SCPOS), text classification with relation and event extraction (TCREE), and an open-domain text classification and NER dataset (TCONER).  
To enable this expansion, we propose an LLM-assisted dataset translation framework that substantially reduces manual annotation cost while preserving annotation consistency.

Based on MMM, we conduct a series of empirical studies to systematically validate the MRE hypothesis. First, we design a unified input-output format and train an Open-domain Information Extraction Large Language Model (OIELLM) on MMM, demonstrating that models trained with MRE mix data achieve stronger and more stable IE performance. Second, we perform controlled ablation experiments across all 21 sub-datasets. The results show that 76\% of the datasets exhibit clear positive reinforcement between word-level and text-level tasks, providing direct empirical evidence of MRE across languages and task types.  
Finally, we incorporate word-level supervision from MRE datasets into a Knowledgeable Verbalizer framework \citep{hu-etal-2022-knowledgeable} for text-level classification. The observed performance gains further support the claim that word-level information encoded by MRE contributes meaningfully to text-level inference.

In summary, this work focuses on empirically validating the Mutual Reinforcement Effect rather than proposing a new model architecture. Our main contributions are as follows:
\begin{enumerate}
    \item We construct the first multilingual MRE mix dataset, MMM, extending existing Japanese-only resources to English and Chinese and covering 21 sub-datasets across multiple IE task combinations.
    \item We provide a systematic empirical validation of the Mutual Reinforcement Effect through large-scale ablation studies across languages, showing that MRE is a robust and reproducible phenomenon rather than a language-specific artifact.
    \item We demonstrate the practical utility of MRE by applying word-level supervision to a Knowledgeable Verbalizer framework, offering additional evidence that MRE enhances text-level IE tasks beyond joint training.
\end{enumerate}

\section{Related Work}

\textbf{Datasets}.  
Existing MRE mix datasets originate from Japanese-only resources, including SCNM \citet{gan2023sentence}, SCPOS \citet{gan2023usa}, and TCREE \citet{gan2023giellm}. While these datasets have demonstrated the feasibility of Mutual Reinforcement Effect, their exclusive focus on Japanese has limited broader empirical validation and cross-lingual exploration of MRE.

In parallel, recent studies have increasingly leveraged Large Language Models (LLMs) for dataset construction and annotation \citep{tan2024large, wadhwa2023revisiting, li-etal-2023-coannotating, laskar2023can}. Prior work has shown that LLM-assisted annotation can substantially reduce human effort while maintaining competitive quality \citep{huang-etal-2023-large}. Such approaches have been applied to diverse domains, including mathematical reasoning datasets \citep{lin2024large} and iterative annotation frameworks such as FreeAL \citep{xiao-etal-2023-freeal}. These methods typically rely on instruction learning and in-context learning to guide LLMs to produce structured labels from unstructured inputs.

Different from prior work, the MMM dataset is designed specifically to support empirical investigation of MRE across languages. By translating existing MRE mix datasets into English and Chinese and expanding them with an open-domain NER dataset (TCONER), MMM enables systematic cross-lingual validation of word-level and text-level task interactions under a unified framework.

\textbf{LLMs for Information Extraction}.  
Generative IE models based on sequence-to-sequence architectures have become increasingly prominent, starting from UIE \citet{lu-etal-2022-unified} and extending to models such as USM \citet{lou2023universal} and Mirror \citep{zhu_mirror_2023}. These approaches unify multiple word-level IE tasks, including NER, relation extraction, and event extraction, through standardized input-output formats, allowing a single model to handle diverse IE subtasks.

With the emergence of LLMs, IE research has largely followed two directions. One line of work directly queries LLMs in zero-shot or few-shot settings using carefully designed prompts \citep{wang2023instructuie, wei2023zero}. Another line focuses on fine-tuning LLMs with task-specific or instruction-based datasets to improve extraction accuracy \citep{zhou2023universalner, xiao2023yayi}. While these studies demonstrate strong IE capabilities, they primarily treat IE subtasks as independent objectives.

In contrast, our work centers on the Mutual Reinforcement Effect, which explicitly studies the bidirectional interaction between word-level and text-level IE tasks within the same input. Rather than proposing a new generative IE architecture, we use MMM as an empirical testbed to examine whether and how such interactions consistently emerge across languages and task combinations.

\section{Multilingual Mutual Reinforcement Effect Mix Datasets}

This section introduces the construction of the Multilingual Mutual Reinforcement Effect Mix (MMM) datasets. Our goal is not to fully automate dataset creation, but to design a practical human--LLM collaborative framework that reduces repetitive manual translation while preserving annotation quality through systematic rule-based filtering and human verification.

\begin{figure}[!h]
\centering
\includegraphics[width=218 pt]{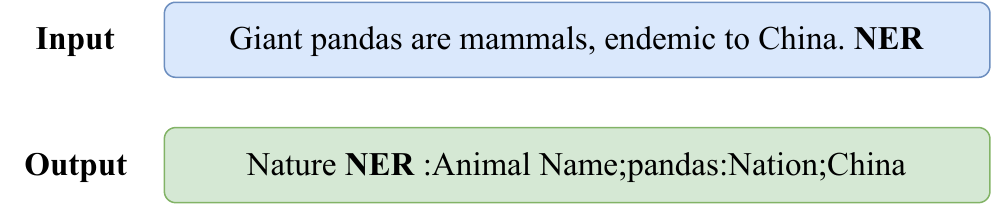}
\caption{\label{4figure4}The format of MMM datasets.}
\end{figure}

\begin{figure*}[!t]
\centering
\includegraphics[width=408 pt]{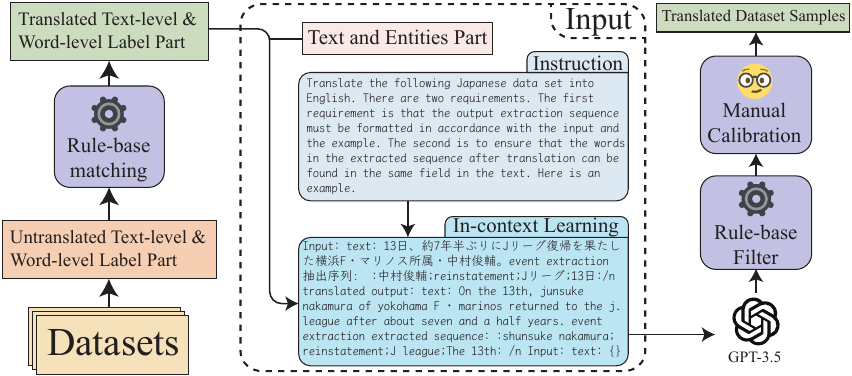}
\caption{\label{3figure3}The overview of the dataset translation framework.}
\end{figure*}

\subsection{Dataset Translation Framework}

We first briefly introduce the MMM dataset format. As shown in Figure~\ref{4figure4}, each sample consists of an input and an output. The input includes the raw text together with a task instruction (e.g., NER), while the output contains a text-level label followed by word-level label--entity pairs. The output format follows the original MRE design, using structured delimiters (e.g., ``:'', ``;'') to ensure consistency across different IE subtasks and languages.

Figure~\ref{3figure3} illustrates the overall translation workflow. We begin by applying rule-based matching to the original Japanese MRE datasets. Since label sets are fixed and shared across datasets, all text-level and word-level labels are translated deterministically using predefined mappings (e.g., ``ポジティブ'' $\rightarrow$ ``positive''). This step is fast, precise, and removes ambiguity before involving LLMs.

The translated labels are then combined with the original text and entity spans and provided to an LLM, GPT-3.5-Turbo \citep{ouyang2022training}, which is used solely to assist in translating the remaining free-text content. We adopt instruction-based prompting together with one-shot in-context learning to guide the model. Instead of optimizing for fully automatic translation accuracy, the prompts explicitly constrain the output format and emphasize entity span consistency with the translated text.

Due to the inherent variability of LLM outputs, we apply a two-stage rule-based filtering process. First, samples containing untranslated Japanese characters are removed. Second, samples with entity spans that cannot be aligned with the translated text or violate MMM formatting constraints are discarded. This filtering step intentionally favors precision over coverage.

The remaining samples are then manually reviewed and calibrated by ten multilingual graduate students proficient in at least two of Chinese, English, and Japanese. Human annotators focus on correcting terminology, resolving minor inconsistencies, and validating domain-specific expressions using external references such as dictionaries. In practice, this framework significantly reduces repetitive translation work, while reserving human effort for quality control rather than raw translation.

Although GPT-3.5-mini serves as the backbone of the pipeline, its role is strictly supportive. Empirically, more advanced models such as GPT-4o did not consistently reduce manual correction effort and were therefore not cost-effective in this setting. Comparative analyses are reported in Appendix~\ref{comparisonbasemodel}. Overall, the proposed framework is best viewed as a scalable annotation aid rather than a fully automated translation system.

\begin{figure*}[!t]
\centering
\includegraphics[width=348 pt]{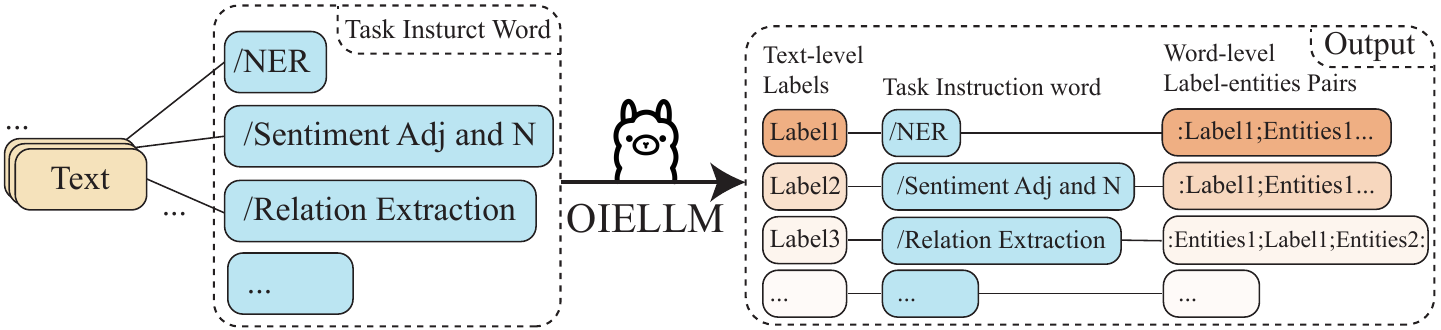}
\caption{\label{5figure5}The input and output of Open-domain Information Extraction Large Language Model (OIELLM).}

\end{figure*}

\section{Open-domain Information Extraction Large Language Model}

This section introduces the Open-domain Information Extraction Large Language Model (OIELLM), which is specifically designed to support empirical investigation of the Mutual Reinforcement Effect on MMM datasets. Rather than proposing a new architecture, OIELLM focuses on a unified input-output formulation that enables large language models to jointly generate text-level labels and word-level label--entity pairs within a single decoding process.

MRE mix datasets differ fundamentally from conventional IE benchmarks. Each sample requires the model to output both a global text-level label and multiple word-level extractions from the same input. Standard sequence labeling models cannot directly support this requirement, while most existing generative IE frameworks primarily target word-level outputs such as entities, relations, or events, without explicitly modeling text-level supervision.

The core objective of OIELLM is to operationalize MRE by enforcing a shared generation space for both levels of supervision. By learning to predict text-level labels and word-level label--entity pairs jointly, the model is encouraged to capture their bidirectional dependency, which forms the basis of MRE. This formulation also improves interpretability, as the generated word-level evidence directly supports the predicted text-level label.

Instead of adopting dialog-style question-answer prompting, we follow prior generative IE paradigms and design a task-agnostic yet structured input-output format tailored to MMM datasets. Figure~\ref{5figure5} illustrates the overall design. Each input consists of the raw text augmented with one or more task instruction words. These instruction words are explicitly marked with a special prefix symbol ``/'' to distinguish them from the text content.

The output follows a fixed and unified structure. It begins with text-level labels, followed by the corresponding word-level label--entity pairs associated with each task instruction. We retain the delimiter-based format (``:'', ``;'') from prior MRE datasets to ensure consistency and unambiguous parsing across tasks and languages.

This format design is critical. By removing dialog prompts and standardizing task specification through instruction tokens, OIELLM reduces input length, avoids prompt-induced variance, and allows the model to focus on learning structural dependencies between text-level and word-level information. Importantly, the same format is shared across all MMM sub-datasets, enabling a single model to process diverse IE tasks under a unified generation framework.

In summary, OIELLM is not merely a fine-tuned LLM, but a structured generation framework that makes MRE observable and measurable across multiple IE subtasks and languages.

\begin{table*}[!ht]
\centering

\resizebox{0.85\textwidth}{!}{

\begin{tabularx}{1.3\textwidth}{@{}l*{9}{>{\centering\arraybackslash}X}@{}}
\toprule[2pt]
Japanese & \multicolumn{3}{c}{SCNM} & \multicolumn{3}{c}{SCPOS: RW} & \multicolumn{3}{c}{SCPOS: Adj \& N} \\
Model  &  TL &  WL &  ALL  & TL &  WL &  ALL& TL &  WL &  ALL\\
\midrule

GPT-3.5-Turbo & 42.07 & 7.54 & 1.97 & 57.20 & 0 & 0 & 28.97 & 5.97 & 0 \\

GPT-4o-mini & 0.27 & 20.61 & 0 & 81.20 & 2.59 & 1.40 & 1.33 & 3.01 & 0 \\

GPT-4o & 58.30 & 23.42 & 8.57 & 87.00 & 4.35 & 1.30 & 82.33 & 0.49 & 0.03 \\

USA-7B & - & - & - & 53.27 & 40.80 & 7.67 & 91.33 & \textbf{81.68} & \textbf{9.63} \\

GIELLM-13B-jp & 85.47 & 84.46 & 54.2 & 86.01 & \textbf{66.61} & \textbf{17.39} & 93.23 & 47.35 & 0.20 \\

OIELLM-8B & 84.73 & 88.53 & 61.93 & 86.50 & 54.76 & 12.40 & 89.13 & 14.88 & 0.40 \\

OIELLM-8B* & 87.30 & \textbf{89.28} & \textbf{64.00} & 88.20 & 53.79 & 12.30 & 89.63 & 15.84 & 0.73 \\

OIELLM-13B & \textbf{89.00} & 86.33 & 57.70 & \textbf{94.60} & 52.36 & 11.90 & \textbf{95.20} & 11.94 & 0.20 \\

\midrule
\end{tabularx}

}


\resizebox{0.85\textwidth}{!}{

\begin{tabularx}{1.3\textwidth}{@{}l*{9}{>{\centering\arraybackslash}X}@{}}
Japanese & \multicolumn{3}{c}{SCPOS: Adj} & \multicolumn{3}{c}{SCPOS: N} & \multicolumn{3}{c}{TCREE} \\
Model &  TL &  WL &  ALL  & TL &  WL &  ALL& TL &  WL &  ALL\\
\midrule

GPT-3.5-Turbo & 65.50 & 0.31 & 0.87 & 39.60 & 6.79 & 0 & 57.20 & 0 & 0 \\

GPT-4o-mini & 0.03 & 0.18 & 0 & 0 & 2.94 & 0 & 0 & 0 & 0 \\

GPT-4o & 68.90 & 0.21 & 0.17 & 74.17 & 0.36 & 0.03 & 90.43 & 0 & 0 \\

USA-7B & 91.43 & 45.51 & 51.77 & 92.03 & \textbf{81.30} & \textbf{9.73} & - & - & - \\

GIELLM-13B-jp & 93.67 & 45.06 & \textbf{55.67} & 92.83 & 46.42 & 0.33 & \textbf{97.47} & \textbf{79.01} & 77.89 \\

OIELLM-8B & 87.13 & 74.96 & 53.07 & 87.77 & 22.92 & 0.50 & 95.07 & 74.92 & 83.69 \\

OIELLM-8B* & 89.93 & \textbf{75.33} & 54.93 & 90.63 & 23.69 & 0.63 & 96.98 & 74.42 & \textbf{84.19} \\

OIELLM-13B & \textbf{94.00} & 60.69 & 42.50 & \textbf{94.70} & 18.07 & 0.60 & 97.08 & 73.82 & 84.19 \\

\bottomrule[2pt]
\end{tabularx}

}

\vspace{1em}


\resizebox{0.85\textwidth}{!}{

\begin{tabularx}{1.3\textwidth}{@{}l*{9}{>{\centering\arraybackslash}X}@{}}
\toprule[2pt]
English & \multicolumn{3}{c}{SCNM} & \multicolumn{3}{c}{SCPOS: RW} & \multicolumn{3}{c}{SCPOS: Adj \& N} \\
Model &  TL &  WL &  ALL  & TL &  WL &  ALL& TL &  WL &  ALL\\
\midrule

GPT-3.5-Turbo & 53.50 & 0.04 & 0 & 14.78 & 2.11 & 0.12 & 68.63 & 13.62 & 0.33 \\

GPT-4o-mini & 0 & 0.03 & 0 & 0 & 0.04 & 0 & 0 & 0 & 0 \\

GPT-4o & 44.63 & 0 & 0 & 86.58 & 3.97 & 0.99 & 85.00 & 17.43 & 0.47 \\

OIELLM-8B & 82.30 & 81.36 & 52.53 & 72.17 & 49.60 & 11.82 & 76.57 & 18.00 & 1.67 \\

OIELLM-8B* & \textbf{85.43} & \textbf{82.38} & \textbf{55.43} & \textbf{74.75} & \textbf{49.93} & \textbf{12.81} & 79.77 & \textbf{19.28} & 2.27 \\

OIELLM-13B & 84.80 & 80.68 & 50.60 & 95.07 & 46.64 & 12.19 & \textbf{94.30} & 18.59 & \textbf{3.20} \\

\midrule
\end{tabularx}

}


\resizebox{0.85\textwidth}{!}{

\begin{tabularx}{1.3\textwidth}{@{}l*{9}{>{\centering\arraybackslash}X}@{}}
English & \multicolumn{3}{c}{SCPOS: Adj} & \multicolumn{3}{c}{SCPOS: N} & \multicolumn{3}{c}{TCREE} \\
Model &  TL &  WL &  ALL  & TL &  WL &  ALL& TL &  WL &  ALL\\
\midrule

GPT-3.5-Turbo & 6.97 & 0.26 & 0.03 & 0.53 & 0.08 & 0 & 12.87 & 0 & 0 \\

GPT-4o-mini & 0 & 0 & 0 & 0 & 0 & 0 & 0 & 0 & 0 \\

GPT-4o & 24.00 & 1.67 & 1.07 & 16.77 & 0.92 & 0.03 & 13.90 & 0 & 0 \\

OIELLM-8B & 75.47 & 51.85 & 32.33 & 76.10 & 28.67 & 1.27 & 80.87 & 21.77 & 33.67 \\

OIELLM-8B* & 76.60 & \textbf{51.95} & 33.17 & 78.67 & 27.45 & \textbf{0.73} & 80.23 & \textbf{25.90} & 22.37 \\

OIELLM-13B & \textbf{94.40} & 50.56 & \textbf{38.40} & \textbf{95.30} & \textbf{28.36} & 0.60 & \textbf{89.90} & 23.50 & \textbf{22.60} \\

\bottomrule[2pt]
\end{tabularx}

}

\vspace{1em}


\resizebox{0.85\textwidth}{!}{

\begin{tabularx}{1.3\textwidth}{@{}l*{9}{>{\centering\arraybackslash}X}@{}}
\toprule[2pt]
Chinese & \multicolumn{3}{c}{SCNM} & \multicolumn{3}{c}{SCPOS: RW} & \multicolumn{3}{c}{SCPOS: Adj \& N} \\
Model &  TL &  WL &  ALL  & TL &  WL &  ALL& TL &  WL &  ALL\\
\midrule

GPT-3.5-Turbo & 41.63 & 9.57 & 2.30 & 50.77 & 2.08 & 0.78 & 59.33 & 7.18 & 0.40 \\

GPT-4o-mini & 5.20 & 18.52 & 0.50 & 12.14 & 7.49 & 0.11 & 0.53 & 1.36 & 0 \\

GPT-4o & 67.43 & 26.11 & 13.47 & 60.38 & 10.91 & 0.99 & 79.27 & 3.69 & 0.50 \\

OIELLM-8B & 84.90 & \textbf{71.90} & 46.40 & 89.29 & 45.75 & 9.93 & 92.33 & 8.75 & 0.33 \\

OIELLM-8B* & 86.33 & 69.97 & \textbf{46.77} & 92.27 & \textbf{46.20} & \textbf{10.60} & 94.50 & \textbf{8.46} & 0.40 \\

OIELLM-13B & \textbf{87.70} & 68.12 & 41.60 & \textbf{95.03} & 43.32 & 8.72 & \textbf{94.90} & 8.42 & \textbf{0.50} \\

\midrule
\end{tabularx}

}


\resizebox{0.85\textwidth}{!}{

\begin{tabularx}{1.3\textwidth}{@{}l*{9}{>{\centering\arraybackslash}X}@{}}
Chinese & \multicolumn{3}{c}{SCPOS: Adj} & \multicolumn{3}{c}{SCPOS: N} & \multicolumn{3}{c}{TCREE} \\
Model &  TL &  WL &  ALL  & TL &  WL &  ALL& TL &  WL &  ALL\\
\midrule

GPT-3.5-Turbo & 56.27 & 0.19 & 0.07 & 53.07 & 3.11 & 0.53 & 59.33 & 7.18 & 0.40 \\

GPT-4o-mini & 27.37 & 1.43 & 0.20 & 5.33 & 1.36 & 0 & 0 & 0 & 0 \\

GPT-4o & 41.60 & 2.00 & 0.83 & 83.93 & 1.45 & 0.47 & 41.60 & 2.00 & 0.83 \\

OIELLM-8B & 93.73 & 60.96 & 53.00 & 92.63 & 28.32 & 0.63 & 91.73 & 58.12 & 56.41 \\

OIELLM-8B* & 95.80 & 64.51 & \textbf{57.63} & 94.97 & \textbf{28.91} & \textbf{1.30} & 95.06 & \textbf{59.54} & \textbf{58.83} \\

OIELLM-13B & \textbf{96.00} & 60.68 & 54.90 & \textbf{95.20} & 27.77 & 1.00 & \textbf{95.26} & 56.91 & 56.00 \\

\bottomrule[2pt]
\end{tabularx}

}

\vspace{1em}


\resizebox{0.85\textwidth}{!}{

\begin{tabularx}{1.3\textwidth}{@{}l*{9}{>{\centering\arraybackslash}X}@{}}
\toprule[2pt]
TCONER & \multicolumn{3}{c}{English} & \multicolumn{3}{c}{Japanese} & \multicolumn{3}{c}{Chinese} \\
Model &  TL &  WL &  ALL  & TL &  WL &  ALL& TL &  WL &  ALL\\
\midrule

GPT-3.5-Turbo & 23.87 & 4.78 & 0 & 23.87 & 2.24 & 0.17 & 29.47 & 2.97 & \textbf{0.57} \\

GPT-4o-mini & 2.93 & 4.06 & 0 & 0 & 3.68 & 0 & 0.03 & 6.12 & 0 \\

GPT-4o & 30.40 & 5.77 & 0.03 & 36.30 & 9.13 & 0.03 & 40.50 & 9.02 & 0 \\

OIELLM-8B & 24.80 & 21.12 & 0.20 & 27.70 & 13.83 & \textbf{0.20} & 33.73 & \textbf{18.87} & 0 \\

OIELLM-8B* & 37.13 & \textbf{23.05} & \textbf{0.30} & 41.40 & \textbf{14.24} & 0.17 & \textbf{48.27} & 18.06 & 0.17 \\

OIELLM-13B & \textbf{40.30} & 19.23 & \textbf{0.30} & \textbf{43.40} & 13.02 & 0 & 47.70 & 15.72 & 0.30 \\

\bottomrule[2pt]
\end{tabularx}

}

\vspace{1em}

\caption{\label{table1results}
\small{The F1 score of MMM datasets. TL F1 score：Text-Level Classification task(e.g. Sentence/Text Classification). WL F1 score: Word-level Label-Entities pairs task(e.g. NER, RE, EE etc.). ALL F1 score: TL and WL are correct simultaneously in one sentence. Note: }}
\end{table*}

\section{Experiment}

This section describes the experimental setup used to empirically evaluate the Mutual Reinforcement Effect, including model training details and evaluation protocols.

\subsection{Details of OIELLM Training}

We select USA-7B (Instruction + In-context Learning)\footnote{\raggedright\url{https://huggingface.co/ganchengguang/USA-7B-instruction-incontext-learning}} and GIELLM-13B-jp\footnote{\raggedright\url{https://huggingface.co/ganchengguang/GIELLM-13B-jpllm}} as baselines, as they are the only prior models explicitly trained on MRE mix datasets. For OIELLM, we adopt Meta-LLaMA3-8B-Instruct\footnote{\raggedright\url{https://huggingface.co/meta-llama/Meta-Llama-3-8B-Instruct}} as the primary backbone. Since LLaMA3 does not provide a 13B variant, we additionally include LLaMA2-13B \citet{touvron2023llama2} for comparison.

\begin{table*}[!t]
\centering

\resizebox{0.60\textwidth}{!}{

\begin{tabular}{lcccccc}
\hline
 English & \textbf{SCNM} & \textbf{SCPOS:RW} & \textbf{SCPOS:adj\&n} & \textbf{SCPOS:adj}& \textbf{SCPOS:n}  & \textbf{TCREE}  \\
\hline
w/o TLI & 80.97 & 48.79 & \textbf{33.29} & 56.04 & \textbf{28.79} & 16.43 \\
with TLI & \textbf{81.28} & \textbf{48.99} & 32.42 & \textbf{56.75} & 27.71 & \textbf{18.43}\\
w/o WLI & 82.40 & 72.41 & 77.27 & 73.73 & 77.07 & 82.23 \\
with WLI & \textbf{83.90} & \textbf{73.15} & \textbf{77.60} & \textbf{75.70} & \textbf{77.73} & \textbf{83.33} \\
\hline
 Chinese & \textbf{SCNM} & \textbf{SCPOS:RW} & \textbf{SCPOS:adj\&n} & \textbf{SCPOS:adj} & \textbf{SCPOS:n}  & \textbf{TCREE}  \\
\hline
w/o TLI & \textbf{73.35} & \textbf{44.36} & 28.67 & 9.68 & 29.06 & 55.10 \\
with TLI & 72.81 & 43.30 & \textbf{29.17} & \textbf{9.73} & \textbf{29.34} & \textbf{56.31} \\
w/o WLI & 83.17 & 89.07 & 91.03 & \textbf{93.67} & 91.80 & 93.64 \\
with WLI & \textbf{83.93} & \textbf{90.95} &  \textbf{92.37} & 92.07 & \textbf{93.63} & \textbf{94.85} \\
\hline
 Japanese & \textbf{SCNM} & \textbf{SCPOS:RW} & \textbf{SCPOS:adj\&n} & \textbf{SCPOS:adj}& \textbf{SCPOS:n}  & \textbf{TCREE}  \\
\hline
w/o TLI & 87.92 & 69.47 & 63.80 & 50.70 & \textbf{67.23} & 80.87 \\
with TLI & \textbf{88.22} & \textbf{69.92} & \textbf{63.89} & \textbf{51.03} & 66.24 & \textbf{81.37} \\
w/o WLI & 83.60 & 87.10 & 88.13 & 87.93 & 88.37 & \textbf{94.86} \\
with WLI & \textbf{85.87} & \textbf{89.50} & \textbf{89.17} & \textbf{89.90} & \textbf{90.57} & 94.46 \\
\hline
 \textbf{TCONER} & English &  & Chinese &  & Japanese &   \\
\hline
w/o TLI & \textbf{20.22} &  & 17.28 &  & 13.19 &  \\
with TLI & 19.85 &  & \textbf{17.82} &  & \textbf{13.39} & \\
w/o WLI & \textbf{36.50} &  & \textbf{44.07} &  & 38.97 &  \\
with WLI & 35.53 &  & 43.33 &  & \textbf{43.30} &  \\

\hline
\end{tabular}
}
\caption{\label{table2MREresults}
The results of text-level information (TLI) and word-level information (WLI) comparison experiments.
}

\end{table*}

We also tested GPT-3.5-Turbo and GPT-4o-mini under one-shot instruction and in-context settings. However, these models failed to consistently generate outputs that conform to the required structured format of MRE mix datasets. Even with sufficient demonstrations, their outputs frequently violated formatting constraints or merged multiple fields, resulting in near-zero F1 scores. As these failures reflect format adherence rather than semantic understanding, we do not include these models as competitive baselines in the main experiments.

OIELLM is fully fine-tuned with all parameters updated. Training is conducted in BF16 precision and inference in FP16. Models are trained for three epochs with a learning rate of $1\times10^{-5}$, using three A800 80GB GPUs and three RTX 6000 Ada 48GB GPUs. Training time ranges from 12 to 20 hours depending on model size. Dataset statistics for training and evaluation splits are reported in Appendix Tables~\ref{table3} and~\ref{table4}.

\subsection{Evaluation}

We adopt F1 score as the primary evaluation metric, following a strict structured prediction protocol. Model outputs are first separated into text-level labels and word-level label--entity pairs according to task instruction words. Word-level outputs are parsed using predefined delimiters (``:'', ``;'') and evaluated as unordered sets.

We report three metrics: Text-Level F1 (TL), Word-Level F1 (WL), and ALL, where ALL measures correctness only when both text-level and word-level outputs are simultaneously correct within a single prediction.

This strict evaluation is intentional. MRE mix datasets are designed as structured information extraction tasks, where outputs must be directly usable as machine-readable annotations. Even minor deviations in formatting, missing fields, or merged entities invalidate the extraction. While general-purpose LLMs often produce semantically reasonable text, they frequently fail to meet these structural constraints, which explains their low scores under this protocol. Detailed evaluation procedures and formulas are provided in Appendix~\ref{appendix:F1score}.

\section{Results}

Table~\ref{table1results} reports the performance of three OIELLM variants trained on the 21 MMM sub-datasets. The model marked with an asterisk, OIELLM-8B, is initialized from LLaMA3-8B-Instruct, while the remaining models use the LLaMA3-8B-Base backbone. Overall, OIELLM achieves strong and stable performance across languages and tasks.

Notably, OIELLM outperforms GIELLM-13B-jp on approximately half of the Japanese datasets, despite GIELLM being specifically designed for Japanese. This result suggests that combining multilingual supervision with MRE-style multitasking can more effectively activate and reuse knowledge embedded in multilingual pretrained LLMs, even for a single target language.

Performance on the TCONER datasets is relatively weaker. We attribute this primarily to data scarcity, as open-domain tasks require substantially larger and more diverse training data than domain-specific settings. Addressing this limitation will be part of future work.

We additionally evaluated GPT-3.5-Turbo and GPT-4o-mini under one-shot instruction and in-context learning settings. Their F1 scores are consistently low under our evaluation protocol. This is mainly due to two factors: (1) the strict structured-output evaluation, where even minor formatting deviations invalidate predictions, and (2) the absence of supervised fine-tuning for MRE-style joint text-level and word-level extraction. These results further motivate the need for dedicated IE models trained explicitly on MRE mix datasets.

\section{Ablation Experiment of MMM Datasets}

Details of the ablation setup are provided in Appendix~\ref{AblationMRE}. Table~\ref{table2MREresults} summarizes the results.

Across the six fixed-label datasets, models trained with additional level information consistently outperform their counterparts trained without it. Overall, \textbf{76\% of the ablation results show that incorporating information from one level facilitates performance at the other level}. This provides strong empirical evidence for the Mutual Reinforcement Effect, confirming that word-level and text-level supervision mutually enhance each other when jointly modeled.

These findings support the central hypothesis of this paper: balanced integration of text-level and word-level tasks improves model understanding and extraction performance, reflecting a reinforcement mechanism analogous to human language comprehension, where local lexical cues and global semantic judgments inform each other.

\begin{table*}[!t]
\centering
\resizebox{0.60\textwidth}{!}{

\begin{tabular}{lcccccc}
\hline
 English & \textbf{SCNM} & \textbf{SCPOS:RW} & \textbf{SCPOS:adj\&n} & \textbf{SCPOS:adj}& \textbf{SCPOS:n}  & \textbf{TCREE}  \\
\hline
Origin KV & 62.95 & 80.42 & 80.40 & 78.87 & 81.95 & \textbf{86.52} \\
WLI KV & \textbf{63.24} & \textbf{83.99} & \textbf{87.40} & \textbf{87.37} & \textbf{88.70} & 85.82 \\
\hline
 Chinese & \textbf{SCNM} & \textbf{SCPOS:RW} & \textbf{SCPOS:adj\&n} & \textbf{SCPOS:adj} & \textbf{SCPOS:n}  & \textbf{TCREE}  \\
\hline
Origin KV & 67.38 & 78.37 & \textbf{91.90} & 84.48 & 84.45 & 93.04 \\
WLI KV & \textbf{71.96} & \textbf{87.97} & 82.92 & \textbf{88.38} & \textbf{87.23} & \textbf{93.95} \\
\hline
 Japanese & \textbf{SCNM} & \textbf{SCPOS:RW} & \textbf{SCPOS:adj\&n} & \textbf{SCPOS:adj}& \textbf{SCPOS:n}  & \textbf{TCREE}  \\
\hline
Origin KV & 73.26 & 30.20 & 67.23 & 73.71 & 73.71 & 73.11 \\
WLI KV & \textbf{73.91} & \textbf{52.90} & \textbf{81.74} & \textbf{85.67} & \textbf{88.31} & \textbf{77.24} \\

\hline
\end{tabular}
}
\caption{\label{table3MREKVresults}
The results of word-level information (WLI) as knowledgeable verbalizer experiments. Compare with original KV construction method. Evaluation task is text classification task.
}

\end{table*}

\begin{figure*}[!h]
\centering
\includegraphics[width=440 pt]{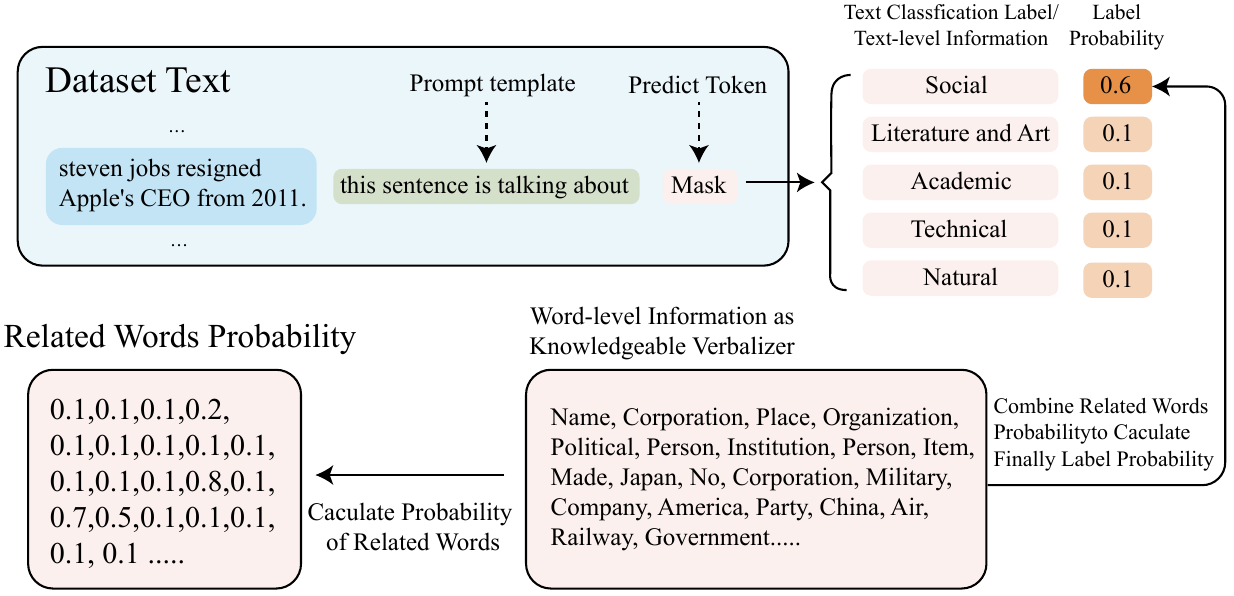}
\caption{\label{Appendix2figure}The figure demonstrates how word-level information is utilized as a Knowledgeable Verbalizer to assist in text-level classification tasks. Additionally, it provides a detailed explanation of the functioning of the Knowledgeable Verbalizer.}

\end{figure*}

For open-domain text classification and NER, the results are more mixed. Some datasets contain multiple or weakly correlated labels, in which not all word-level information contributes positively to text-level prediction. This reduces the overall reinforcement effect. However, for the Chinese and Japanese TCONER datasets, incorporating level information consistently improves performance. This observation suggests that MRE may be more effective in character-based languages, where lexical units often carry richer semantic information, compared to alphabetic languages such as English.

These ablation results establish a clear empirical foundation for further analysis of how word-level information can be selectively leveraged to enhance text-level inference, which we explore next through Knowledgeable Verbalizer experiments.

\section{Word-level Information as Knowledgeable Verbalizer}

To examine whether the Mutual Reinforcement Effect (MRE) extends beyond information extraction, we conduct an additional empirical study on few-shot text classification. In the MRE framework, word-level information (WLI) is expected to support text-level understanding. Accordingly, we use WLI extracted from MRE mix datasets to construct Knowledgeable Verbalizers (KV) \citep{hu-etal-2022-knowledgeable}.

For each classification label, we select the top-$100$ high-frequency words from the corresponding word-level annotations as label-specific verbalizers. During inference, token probabilities are aggregated at the label level, and the label with the highest aggregated probability is predicted. Compared with conventional KV construction based on general-purpose related-word resources, WLI-based KVs are task-aligned and dataset-specific. Superior performance of WLI-based KVs thus provides indirect evidence that word-level information facilitates text-level classification, supporting the MRE hypothesis beyond IE tasks.



For the empirical experiments on fine-tuning, we selected the LLaMA3-8B\footnote{https://ai.meta.com/blog/meta-llama-3/} model\footnote{https://huggingface.co/meta-llama/Meta-Llama-3-8B} as the base model to perform a series of fine-tuning and inference tasks. We opted not to use the LLaMA3-8B-Instruct version because it is more tailored for question-answering tasks, with prompts structured as instructions. Through a comparative analysis of LLaMA3-8B and its instruct-tuned counterpart, we observed that the base LLaMA3-8B model achieved better performance on fundamental IE tasks. Therefore, we decided to use LLaMA3-8B as the foundation for our experiments.

For the WLI as KV application comparison experiments, we employed the T5-base \citet{raffel2020exploring} model as the base model. Specifically, for the English portion of the MMM dataset, we used the original Google T5-base\footnote{https://huggingface.co/google-t5/t5-base}. For the Chinese section, we selected the Mengzi-T5-base\footnote{https://huggingface.co/Langboat/mengzi-t5-base}, which is optimized for Chinese tasks. Lastly, for the Japanese part of the MMM dataset, we utilized T5-base-Japanese\footnote{https://huggingface.co/sonoisa/t5-base-japanese}.

For the fine-tuning experiment, the entire training set was utilized to fully parameterize the fine-tuned LLMs. Subsequently, 1,000 samples were randomly selected from the test set three times, and the results from these three trials were averaged to produce the final performance score. The evaluation metric employed was the F1 score.

The hyperparameters for training were configured as follows: the number of training epochs was set to 3, and the learning rate was initialized at 1e-5. The AdamW optimizer was used, with 100 warm-up steps. Training was conducted on three RTX A6000 Ada GPUs, each with 48 GB of memory. To optimize GPU memory usage, BF16 precision was applied during training, and FP16 precision was employed for inference.

\subsection{Results of Word-level Information as Knowledgeable Verbalizer}

Table~\ref{table3MREKVresults} shows KV-based text classification results on 18 MMM sub-datasets in English, Chinese, and Japanese. The open-domain TCONER dataset is excluded due to its unfixed label schema.

WLI-based KVs achieve the best performance on 16 out of 18 datasets and consistently outperform the original KV construction method. Gains are especially notable on sentiment classification tasks, where word-level polarity is crucial. Unlike the ablation experiments that evaluate MRE within IE tasks, these results demonstrate that WLI learned through MRE transfers effectively to downstream text classification, providing additional empirical evidence for the existence and practical utility of MRE.

\section{Conclusion and Future Work}

This paper presents a large-scale empirical study of the MRE in information extraction. We construct the MMM datasets by translating existing Japanese MRE datasets into English and Chinese and by introducing the TCONER dataset to cover open-domain IE tasks. To reduce annotation cost, we adopt an LLM-assisted translation framework that supports, rather than replaces, human annotation.

Based on the MMM datasets, we train an OIELLM with a unified input-output format that jointly models text-level and word-level information. Extensive experiments show that OIELLM achieves strong multilingual IE performance. More importantly, ablation studies on 21 sub-datasets demonstrate that in 76\% of cases, information at one level facilitates learning at the other level, providing direct empirical evidence for the MRE hypothesis.

We further apply word-level information as Knowledgeable Verbalizers in few-shot text classification and observe consistent improvements over conventional KV construction methods. These results indicate that MRE-derived word-level information can effectively enhance downstream text-level tasks, forming a threefold empirical validation of MRE across ablation analysis, IE modeling, and downstream application.

Future work will extend MRE to more languages and task combinations and explore incorporating MRE-aware objectives into pre-training and instruction tuning of large language models.

\section{Limitations}

This study has several limitations worth noting. First, although we experimented with different large language models for dataset translation, our empirical results show that GPT-4o does not consistently outperform GPT-3.5-mini in the context of MRE dataset translation. Due to budget constraints, we did not exhaustively evaluate all available models across all sub-datasets.

Second, the proposed dataset translation framework is not designed to fully automate dataset construction. Instead, it functions as an annotation-assistance tool that reduces repetitive translation effort for relatively simple cases, while leaving complex or ambiguous instances to human annotators. All translated samples are therefore manually verified and refined. As a result, we do not report automatic translation quality metrics, and instead provide the proportion of manually corrected samples as a coarse indicator of translation reliability.

Finally, while our empirical results demonstrate the Mutual Reinforcement Effect across multiple tasks and languages, the current study focuses on specific combinations of text-level and word-level information. Extending MRE to other task configurations and exploring automatic criteria for identifying beneficial task interactions remain open directions for future work.

\section*{Acknowledgments}

This research was supported in part by JSPS KAKENHI Grant Numbers JP24K15084 and JP23H00491.

\bibliography{custom}

\appendix


\begin{figure*}[!t]
\centering
\includegraphics[width=440 pt]{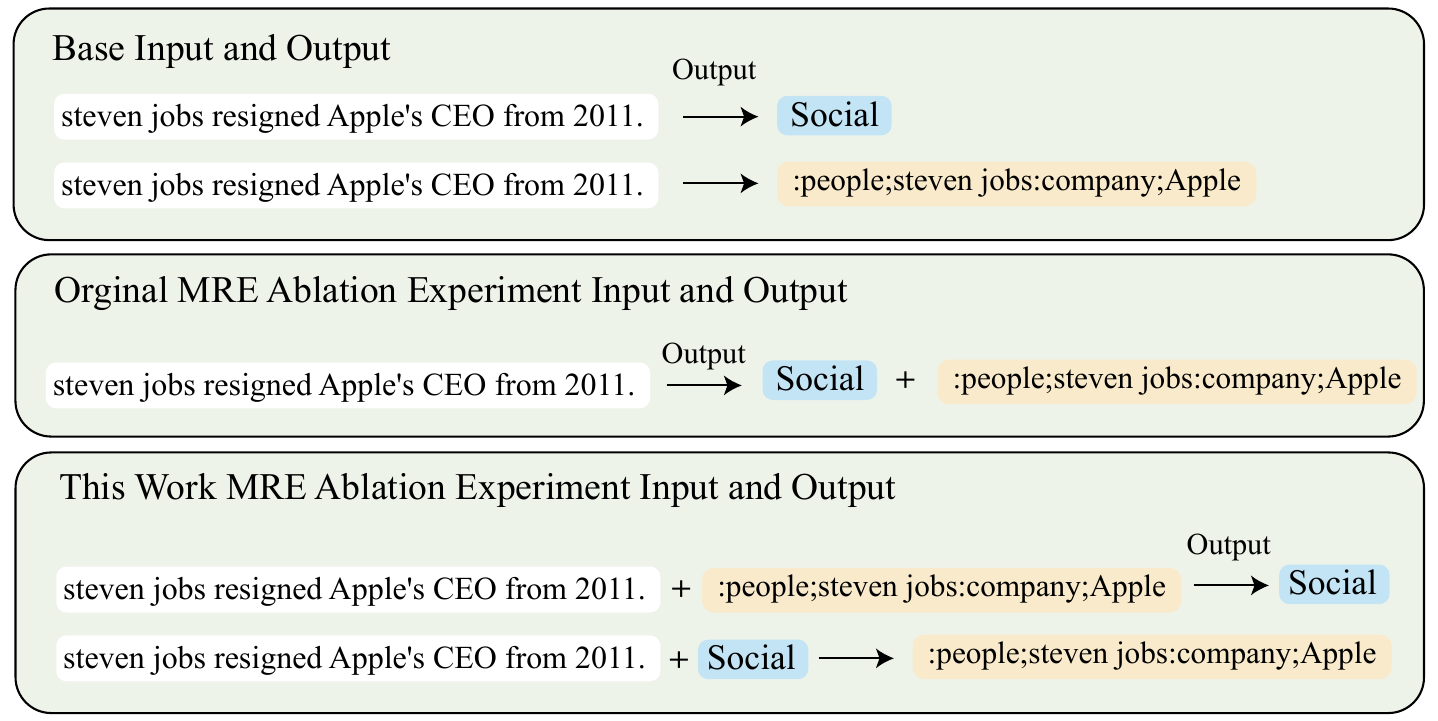}
\caption{\label{Apeendix2DMRE}The figure shows the inputs and outputs of the traditional ablation experiment for the MRE task and the new empirical MRE experiment proposed in this work.}

\end{figure*}

\section{Empirical Experiment of Mutual Reinforcement Effect}\label{AblationMRE}

The three format of fine-tuned language models used for ablation experiments are shown in Figure \ref{Apeendix2DMRE}. The sentence on the left represents the input, with the plus sign indicating the addition of Word-level Information (WLI. i.e. Word-level Task) or Text-level Information (TLI. i.e. Text-level Task), which are appended to the sentence to form the full input. The arrows represent the output produced by language model. The distinctions between the models are clearly illustrated.

First, the top model in Figure \ref{Apeendix2DMRE} shows the input-output format for the traditional IE task, where language models are fine-tuned on a basic input sentence. The model then outputs either classified labels or extracted label-entity pairs. This approach treats the two tasks—word-level label extraction and text-level classification—independently, with no shared information between them.

In contrast, the middle section of Figure \ref{Apeendix2DMRE} illustrates the input-output format for the original MRE task. While the input remains a single sentence, the model is expected to output both word-level label-entity pairs and text-level classification labels simultaneously. Thus, during MRE fine-tuning, the model learns to capture both levels of information, integrating the two tasks.

\begin{figure}[!h]
\centering
\includegraphics[width=219 pt]{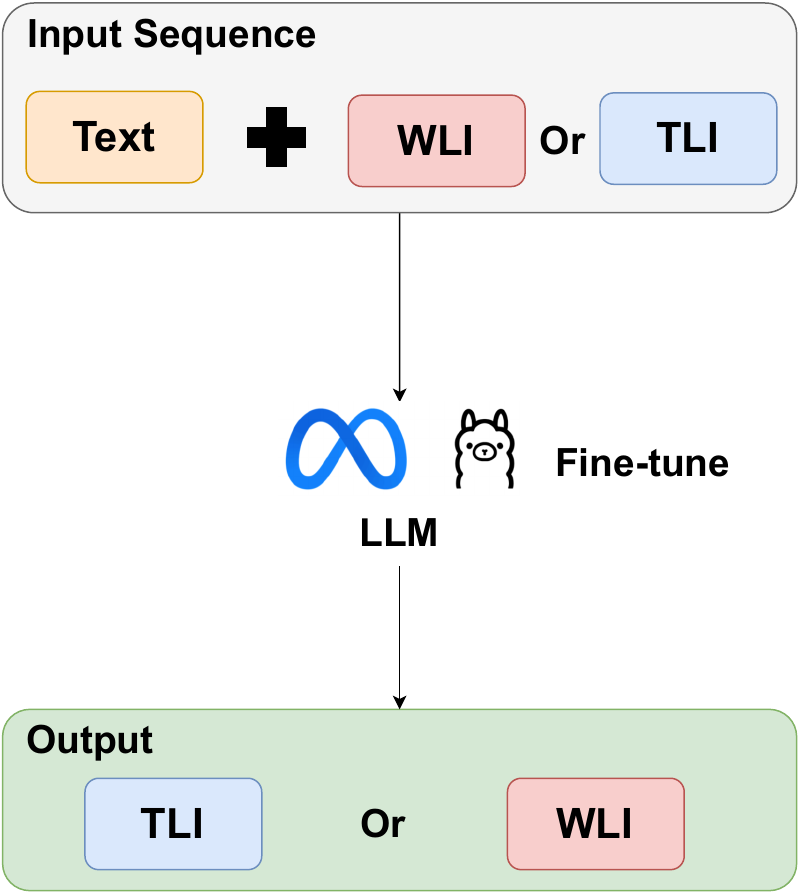}
\caption{\label{Appendix3figure}The figure illustrates the flow of an empirical MRE experiment using the new approach.}

\end{figure}

Finally, the bottom section of Figure \ref{Apeendix2DMRE} presents the input-output format of our proposed ablation experiment designed to validate MRE. Unlike the previous two formats, this approach aims to verify the existence of shared knowledge between word-level and text-level tasks. Specifically, we introduce WLI and TLI to both levels of tasks to assess whether enhancing one task also improves the other. For example, by adding word-level label-entity pairs to the input text and asking the model to output the text-level classification label, we can evaluate whether the additional word-level information assists in text classification. Similarly, if adding text-level information to the input improves the extraction of word-level label-entity pairs, it suggests the presence of an MRE between the two tasks.

As showed in Figure \ref{Appendix3figure}, the LLM is fine-tuned with all parameters using revised input and output formats. The input sequence is directly concatenated with either WLI or TLI, while the output consists solely of TLI or WLI. No additional instruction templates or prompt words were incorporated in this process. We deliberately concatenated the text with WLI or TLI without extra modifications to minimize the potential influence of extraneous words or sentences on the model’s output, which could affect the accuracy of our comparative experiments. By using only this basic spliced input and raw output, we aim to investigate whether tasks at one level facilitate tasks at another, while controlling for other confounding factors.

To test this hypothesis, we conducted ablation experiments on 21 sub-datasets of Multilingual MRE Mix (MMM) datasets. The results were analyzed to further deepen our understanding of MRE and its implications.

Second, for the experiments involving the knowledgeable verbalizer, we utilized the OpenPrompt\citet{ding2021openprompt}\footnote{https://github.com/thunlp/OpenPrompt} framework to efficiently set up the experimental environment. All datasets were divided into training and test sets. From the training set, we randomly selected 20 samples per category, based on the label types, to form the prompt experiment’s training subset. Each experiment was trained for 2 epochs, with all other hyperparameters—such as the learning rate—kept consistent across experiments. The only variation lay in the construction method of the KV.

For the KVs based on the original approach, we leveraged ChatGPT-4o\footnote{https://chatgpt.com/} to generate the top 100 most relevant words for each label. In contrast, for KVs constructed using the WLI-based method, we developed a custom processing script. The script segmented all words from the WLI section of each dataset, identified high-frequency terms, and used them to construct the WLI-based KVs.

\section{Construction of TCONER}\label{appendix:MMMdataset}

In the original MRE mix datasets, relation and event extraction tasks are open-domain, implying that the labels are not predefined. However, the label set is limited to only a dozen options. Given this context, we constructed a new dataset, termed TCONER, based on an open-domain Named Entity Recognition (NER) dataset\footnote{\raggedright\url{https://huggingface.co/datasets/Universal-NER/Pile-NER-type?row=0}} \citep{zhou2023universalner}. The labels at the text level in the TCONER dataset are also open-domain. To annotate this dataset, we initially employed the GPT-3.5-Turbo model to assign open-domain text-level labels. Subsequent manual verification and annotation were conducted to ensure accuracy and consistency, resulting in the finalized TCONER dataset. Similarly, we translated the constructed English TCONER dataset using the dataset translation framework. The TCONER dataset was translated into Japanese and Chinese.



\begin{table}[!h]
\centering
\begin{tabular}{lccc}
\hline
  \textbf{Dataset}  & \textbf{SCNM} & \textbf{SCPOS: RW} &  \textbf{SCPOS:}  \\
  &  & \textbf{RW} & \textbf{Adj \& N} \\
\hline
Japanese & \textbf{5343} & \textbf{2000} & \textbf{187528} \\
English & 4449 & 1312 & 4801 \\
Chinese & 3177 & 1406 & 3937 \\

\hline
\end{tabular}

\vspace{0.3cm}

\begin{tabular}{lccc}
\hline
  \textbf{Dataset}  & \textbf{SCPOS:} & \textbf{SCPOS:} &  \textbf{TCREE}  \\
  & \textbf{Adj} & \textbf{N} &  \\
\hline
Japanese & \textbf{187528} & \textbf{187528} & \textbf{2000} \\
English & 9132 & 5027 & 1910 \\
Chinese & 7413 & 3920 & 1491 \\

\hline
\end{tabular}

\vspace{0.3cm}

\begin{tabular}{lccc}
\hline
  \textbf{Language}  & \textbf{English} & \textbf{Japanese}   & \textbf{Chinese}  \\

\hline
TCONER & \textbf{45888} & 6791  & 9047 \\

\hline
\end{tabular}

\caption{\label{table2}
 Statistical results of the translated MMM dataset. (Due to resource constraints, we extracted only 10,000 samples as translation objects from each of the three SCPOS sub-datasets and the TCONER dataset.)}
\end{table}

Table \ref{table2} presents the statistics of the final translation results. Due to the high costs associated with the use of a premium API, we limited our study to 10,000 samples from each of three sub-datasets within SCPOS and the TCONER dataset, which contains 180,000 entries. These 10,000 samples, retained post-translation, proved to be an ample test set. It was observed that there was a greater data loss when translating into Chinese compared to English. This discrepancy may be attributed to the training data predominance of English in OpenAI's GPT-3.5-Turbo model, resulting in superior performance in English-related tasks. For instance, in the SCNM and TCREE datasets, the Japanese to English translation accuracy exceeded 80\%. Conversely, the translation results from English to Chinese in the TCONER dataset were markedly better than those from English to Japanese. This further confirms that GPT-3.5-Turbo exhibits enhanced effectiveness with major languages compared to lesser-used ones.

\begin{table}[!h]
\centering
\begin{tabular}{lccc}
\hline
  \textbf{Dataset}  & \textbf{SCNM} & \textbf{SCPOS: RW} &  \textbf{SCPOS:}  \\
  &  & \textbf{RW} & \textbf{Adj \& N} \\
\hline
Japanese & 1000 & 1000 & 1000 \\
English & 1000 & 500 & 1000 \\
Chinese & 1000 & 500 & 1000 \\

\hline
\end{tabular}

\vspace{0.3cm}

\begin{tabular}{lccc}
\hline
  \textbf{Dataset}  & \textbf{SCPOS:} & \textbf{SCPOS:} &  \textbf{TCREE}  \\
  & \textbf{Adj} & \textbf{N} &  \\
\hline
Japanese & 1000 & 1000 & 1000 \\
English & 1000 & 1000 & 500 \\
Chinese & 1000 & 1000 & 500 \\

\hline
\end{tabular}

\vspace{0.3cm}

\begin{tabular}{lccc}
\hline
  \textbf{Language}  & \textbf{English} & \textbf{Japanese}   & \textbf{Chinese}  \\

\hline
TCONER & 2000 & 2000  & 2000 \\

\hline
\end{tabular}

\caption{\label{table3}
 Statistical results of train sets of OIELLM.}
\end{table}

\begin{table}[!h]
\centering
\begin{tabular}{lccc}
\hline
  \textbf{Dataset}  & \textbf{SCNM} & \textbf{SCPOS: RW} &  \textbf{SCPOS:}  \\
  &  & \textbf{RW} & \textbf{Adj \& N} \\
\hline
Japanese & 4343 & 1000 & 186528 \\
English & 3449 & 812 & 3801 \\
Chinese & 2177 & 906 & 2937 \\

\hline
\end{tabular}

\vspace{0.3cm}

\begin{tabular}{lccc}
\hline
  \textbf{Dataset}  & \textbf{SCPOS:} & \textbf{SCPOS:} &  \textbf{TCREE}  \\
  & \textbf{Adj} & \textbf{N} &  \\
\hline
Japanese & 186528 & 186528 & 1000 \\
English & 8132 & 4027 & 1410 \\
Chinese & 6413 & 2920 & 991 \\

\hline
\end{tabular}

\vspace{0.3cm}

\begin{tabular}{lccc}
\hline
  \textbf{Language}  & \textbf{English} & \textbf{Japanese}   & \textbf{Chinese}  \\

\hline
TCONER & 43888 & 4791  & 7047 \\

\hline
\end{tabular}

\caption{\label{table4}
 Statistical results of test sets.}
\end{table}

\section{Statistical Results of Train and Test Dataset in OIELLM}
\label{appendix:TrainTestdataset}

As shown in Tables \ref{table3} and \ref{table4}, the statistics for the complete training and test sets of the MMM dataset. The MMM dataset was segmented into 21 sub-datasets. Training set sizes were assigned based on the sizes of these sub-datasets, categorized into three groups: 500, 1000, and 2000 samples. Samples beyond these numbers were allocated to the test sets.

\begin{table*}[!t]
\centering

\begin{tabular}{|l|c|c|c|c|}
\hline
\textbf{Dataset} & \textbf{Direction} & \textbf{Model} & \textbf{Remaining Samples} & \textbf{Lost \%} \\
\hline
SCNM & JA $\rightarrow$ EN & GPT-3.5-Turbo & 4449 & 16.73\% \\
SCNM & JA $\rightarrow$ EN & GPT-4o         & 3996 & 25.21\% \\
\hline
SCPOS: RW & JA $\rightarrow$ EN & GPT-3.5-Turbo & 1312 & 34.40\% \\
SCPOS: RW & JA $\rightarrow$ EN & GPT-4o         & 1182 & 41.04\% \\
\hline
SCPOS: Adj\&N & JA $\rightarrow$ EN & GPT-3.5-Turbo & 4801 & 51.99\% \\
SCPOS: Adj\&N & JA $\rightarrow$ EN & GPT-4o         & 4821 & 51.79\% \\
\hline
TCREE & JA $\rightarrow$ EN & GPT-3.5-Turbo & 1910 & 4.50\% \\
TCREE & JA $\rightarrow$ EN & GPT-4o         & 1979 & 1.05\% \\
\hline
TCONER & EN $\rightarrow$ ZH & GPT-3.5-Turbo & 9047 & 9.53\% \\
TCONER & EN $\rightarrow$ ZH & GPT-4o         & 7951 & 20.49\% \\
\hline

\end{tabular}
\caption{Translation Quality Comparison between GPT-3.5-Turbo and GPT-4o}\label{comparetranslationmodel}

\end{table*}

\begin{table*}[!t]
\centering
\begin{tabular}{lll}
\hline
 Datasets & \textbf{Text-level} & \textbf{Word-level}    \\
\hline
 SCNM & Society, Literature,  & people, corporations, political \\
  & Academia, Technology, &   organizations, other organizations, \\
& Nature  &   places, facilities, products, and events     \\
\hline
 SCPOS:RW& positive, negative & positive, neutral, negative \\
 \hline
 SCPOS:N& positive, negative & positive, neutral, negative   \\
 \hline
 SCPOS:Adj& positive, negative & positive, negative  \\
 \hline
 SCPOS:N \& Adj& positive, negative & positive, neutral, negative  \\
 \hline
 TCREE & sports, film, women,  & affiliation, occupation, starring, director,  \\
  & IT, advertising  &  age, product, goods, performances, wins,    \\
  &   &   broadcasts, public appearances, launches,  \\
  &   &   retirements  \\
  \hline
  TCONER & Entertainment, Politics & date, location, organization   \\
         & Medical, Health, education  & Title, Person, City  \\
         & Tech, Healthcare, News  & Law, Number, Concept    \\
         & finance, Biolog, etc.  &  TV Show, Object, etc.  \\
\hline
\end{tabular}
\caption{\label{table1mixdataset}
The table presents seven distinct types of MRE mixed datasets, each available in Chinese, English, and Japanese, resulting in a total of 21 sub-datasets. Among them, the TCONER dataset corresponds to an open-domain dataset, where only a subset of the labels is provided, rather than a comprehensive list of all possible labels. (SCNM: Sentence Classification and Named Entity Recognition Mix Dataset. SCPOS: Sentiment Classification and Part-of-Speech Dataset. RW: Relation Word. N: Noun. Adj: Adjective. N \& Adj: Nouns and Adjective. TCREE: Text Classification and Relation \& Event Extraction Dataset. TCONER: Open-domain Text Classification and NER mix dataset)}
\end{table*}

\begin{figure*}[ht]
\centering
\includegraphics[width=428 pt]{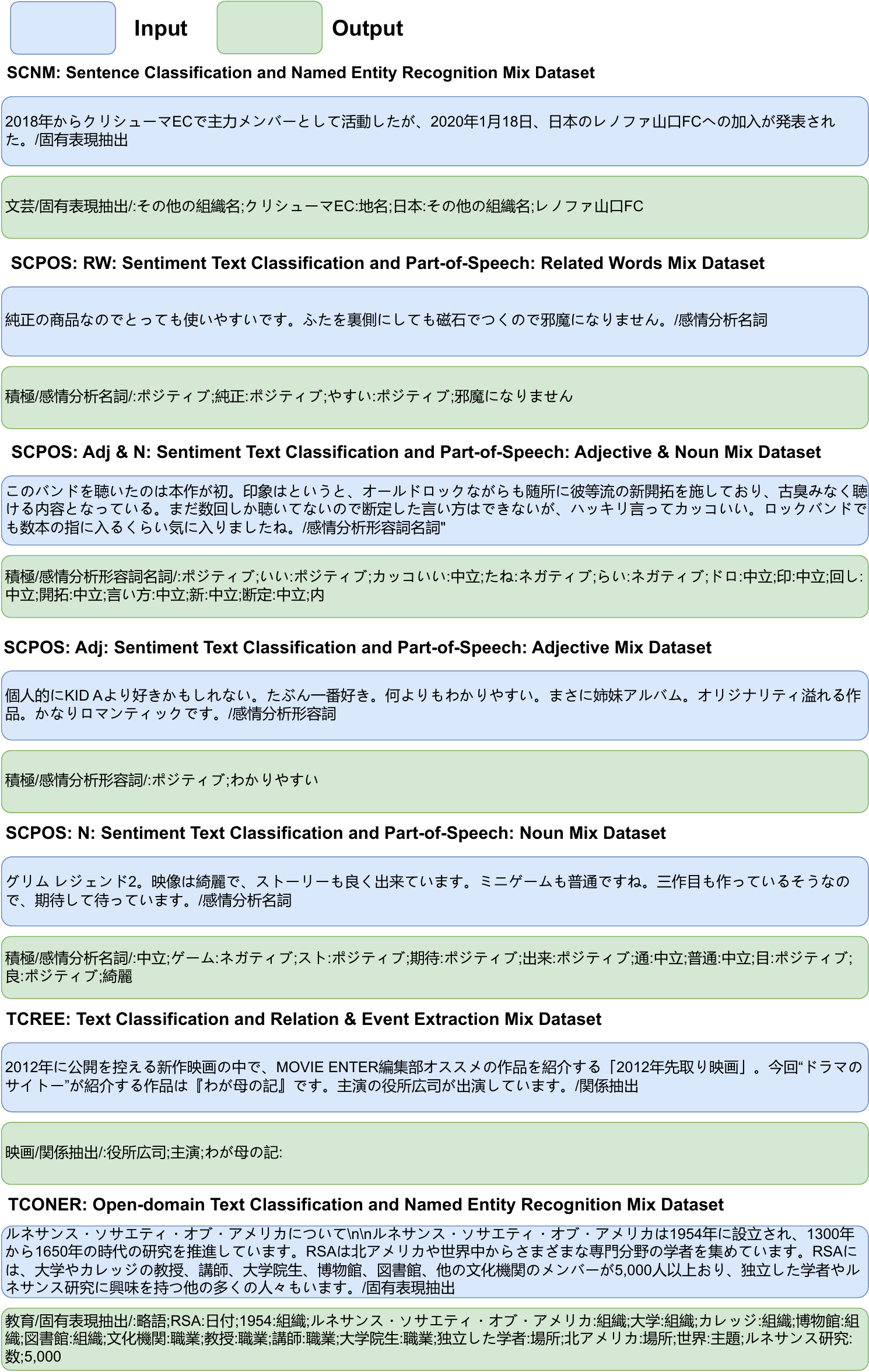}
\caption{\label{6figure6}The input and output format example with OIELLM in Japanese MRE mix datasets.}

\end{figure*}

\begin{figure*}[ht]
\centering
\includegraphics[width=428 pt]{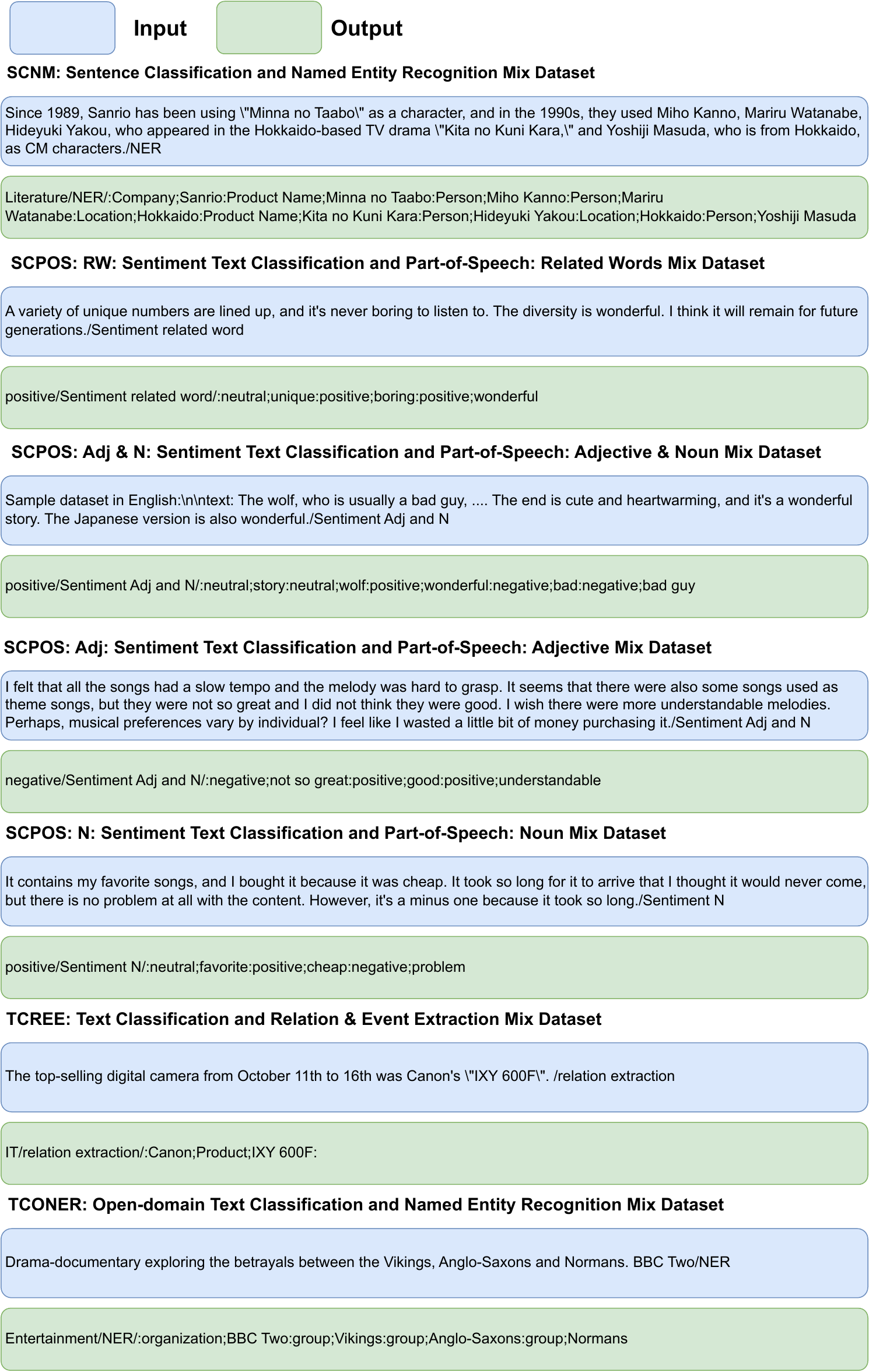}
\caption{\label{7figure7}The input and output format example with OIELLM in English MRE mix datasets.}

\end{figure*}

\begin{figure*}[ht]
\centering
\includegraphics[width=428 pt]{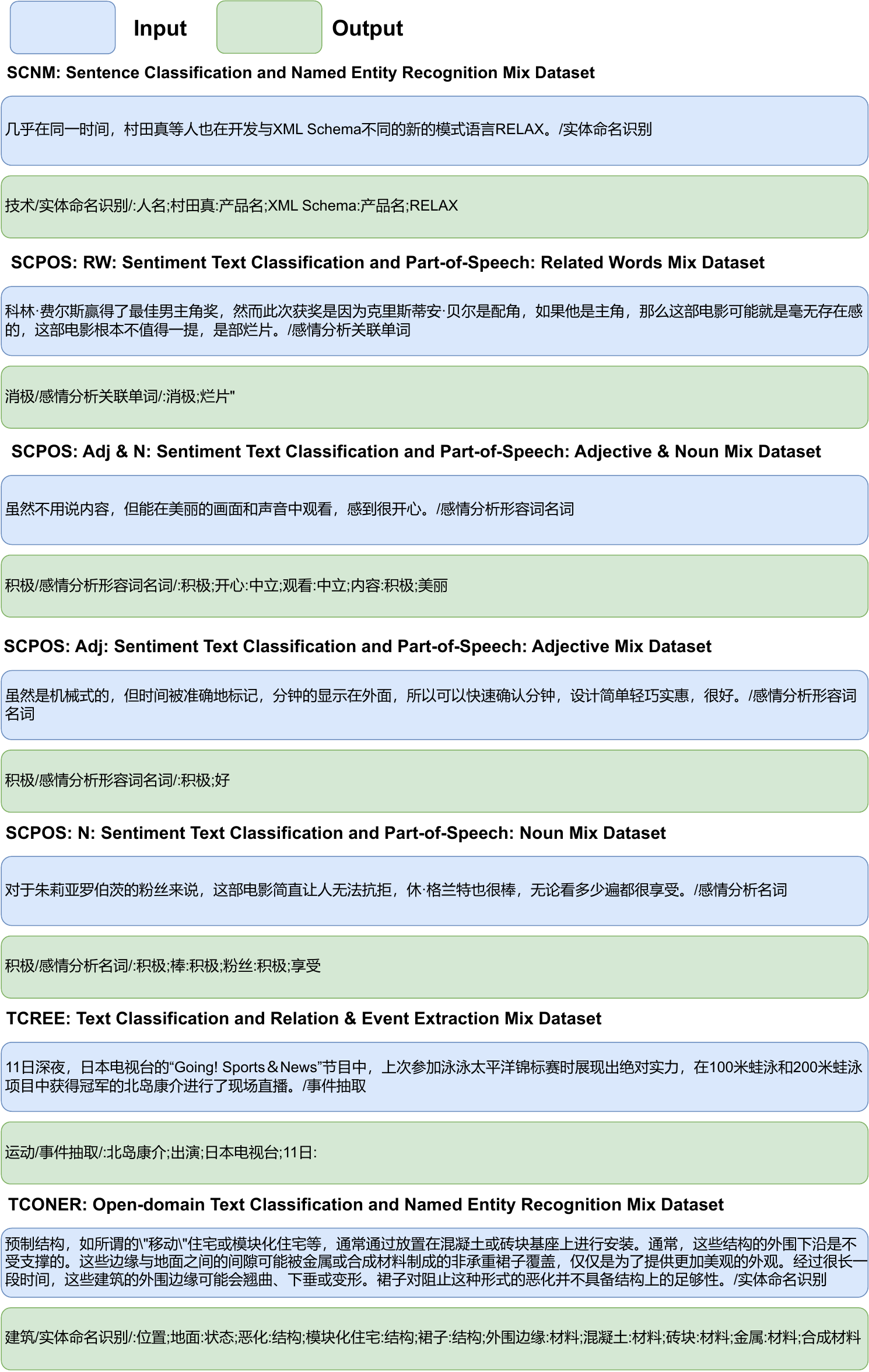}
\caption{\label{8figure8}The input and output format example with OIELLM in Chinese MRE mix datasets.}

\end{figure*}

\section{Comparison of GPT-4o and GPT-3.5-mini on Dataset Translation Frameworks}\label{comparisonbasemodel}

In our translation pipeline, we initially selected GPT-3.5-Turbo (referred to as GPT-3.5-mini) for translating Japanese datasets due to its favorable trade-off between translation quality and cost-efficiency. Although GPT-4o is a more advanced model, our empirical tests suggest that it did not consistently outperform GPT-3.5-mini in our use case, particularly for domain-specific and structured datasets relevant to Mutual Reinforcement Effect (MRE) tasks.

To validate this decision, we conducted a comparative evaluation of both models across multiple datasets. The workflow included two rounds of rule-based filtering to remove untranslated Japanese text and span mismatches, followed by two rounds of manual review. We report the proportion of samples retained after rule-based filtering, which serves as an indirect indicator of translation consistency and quality. The results are summarized in Table~\ref{comparetranslationmodel}.

As shown, GPT-4o exhibited slight improvements in a few datasets (e.g., TCREE) but also underperformed in others (e.g., SCNM and TCONER), resulting in a higher sample loss percentage after filtering. These results support our initial choice of GPT-3.5-Turbo as a more cost-effective and stable translation backbone.

Despite this, we recognize that future improvements in translation models could further reduce manual correction efforts. For instance, although our current framework already reduced 20--40\% of samples through rule-based filtering and required manual correction in only ~10\% of the remaining data, employing more accurate models may streamline this process even further. We plan to incorporate GPT-4o or its successors in subsequent updates of the MMM dataset.

\section{UI of the Tools for Manually Correcting Translated Dataset}\label{annotationtools}

\begin{figure*}[ht]
\centering
\includegraphics[width=428 pt]{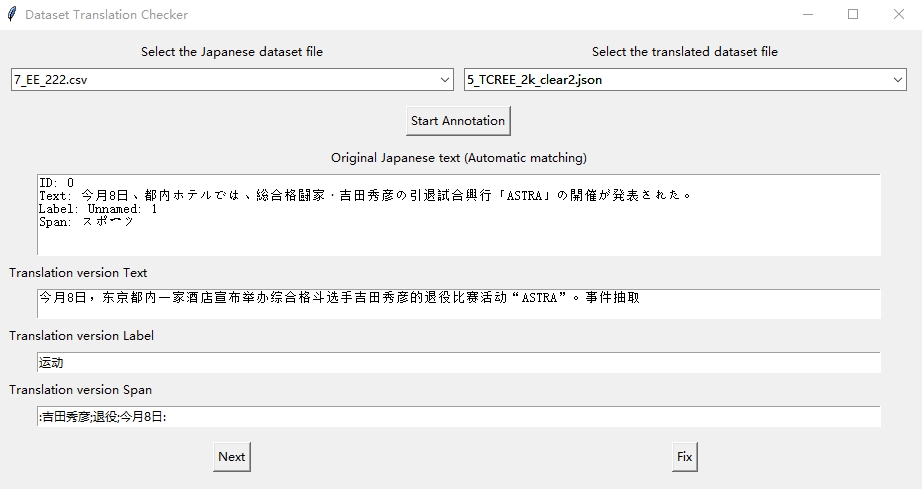}
\caption{\label{9figure9}UI of the Tools for Manually Correcting Translated Dataset.}

\end{figure*}

As illustrated in the figure \ref{9figure9}, the top section of the UI interface allows users to select two dataset files: the original Japanese dataset and its translated counterpart in either English or Chinese. Once selected, corresponding samples from both the original and translated datasets are displayed below for direct comparison. Users can manually edit the translations as needed and apply corrections by clicking the “Fix” button. If the translation is accurate, the user can proceed to the next sample by clicking “Next,” streamlining the verification process.

\section{Calculate Detail of F1 Score}\label{appendix:F1score}

\begin{equation}
    F_1 = 2 \times \frac{\text{precision} \times \text{recall}}{\text{precision} + \text{recall}}
\end{equation}
\begin{equation}
    \text{precision} = \frac{|Real \cap Generated|}{|Generated|}
\end{equation}
\begin{equation}
    \text{recall} = \frac{|Real \cap Generated|}{|Real|}
\end{equation}

\begin{algorithm}
\caption{Parse Text Label and Entity Pairs}
\begin{algorithmic}[1]
\Procedure{parse\_output}{output, instruct\_word, is\_tcree}
\State \textbf{Input:} \textit{output} (String), \textit{instruct\_word} (String), \textit{is\_tcree} (Boolean)
\State \textbf{Output:} \textit{text\_label} (String), \textit{entity\_pairs} (Set of Tuples)
\State
\State $instruct\_word \gets$ \textit{instruct\_word}
\If{$instruct\_word \notin output$}
    \State \Return $(" ", \{\})$
\EndIf
\State $text\_label, entity\_pairs \gets$ \textit{output.split(instruct\_word, 1)}
\State $text\_label \gets text\_label.strip()$
\If{\textit{is\_tcree}}
    \State $entity\_pairs \gets [entity\_pairs.strip()]$
\Else
    \State $entity\_pairs \gets [pair.strip()$ \textbf{for} $pair$ \textbf{in} $entity\_pairs.split(":")$ \textbf{if} $pair]$
\EndIf
\State $entity\_pairs \gets [tuple(pair.split(";"))$ \textbf{for} $pair$ \textbf{in} $entity\_pairs]$
\State \Return $(text\_label, \text{set}(entity\_pairs))$
\EndProcedure
\end{algorithmic}
\end{algorithm}

\section{Case Study of Input and Output Format with OIELLM in MRE mix datasets}

\end{CJK}
\end{document}